\documentclass[sigconf]{acmart}

\setcopyright{none}
\renewcommand\footnotetextcopyrightpermission[1]{}

\usepackage{amsmath}
\usepackage{graphicx}
\usepackage{multirow}
\usepackage{booktabs}
\usepackage{xcolor}
\usepackage[most]{tcolorbox}

\newtcolorbox{promptbox}[1]{
    colback=gray!5,
    colframe=black!65,
    fonttitle=\bfseries,
    title=#1,
    boxrule=0.6pt,
    arc=2pt,
    left=5pt,
    right=5pt,
    top=5pt,
    bottom=5pt,
    breakable
}

\title{Detecting Differences Is Not Understanding Structure: Large Language
Models Fail at Graph Isomorphism}

\author{Kumar Thushalika}
\affiliation{%
  \institution{University of Ruhuna}
  \city{Galle}
  \country{Sri Lanka}
}
\email{thushalika\_k\_e23@engug.ruh.ac.lk}

\author{Sukumar Kishanthan}
\affiliation{%
  \institution{University of Moratuwa}
  \city{Moratuwa}
  \country{Sri Lanka}
}
\email{sukumar@cse.mrt.ac.lk}

\author{Asela Hevapathige}
\affiliation{%
  \institution{University of Melbourne}
  \city{Melbourne}
  \country{Australia}
}
\email{asela.hevapathige@unimelb.edu.au}

\acmConference[GaLM @ CIKM 2026]
{GaLM Workshop at CIKM 2026}
{November 8, 2026}
{Rome, Italy}

\begin{document}

\begin{abstract}
Large language models (LLMs) have shown impressive performance on diverse reasoning tasks, yet their capacity for structural reasoning in graphs remains unclear. We investigate whether LLMs can genuinely understand graph isomorphism—a fundamental problem in graph theory. In our experiments, the three evaluated LLMs achieved nearly perfect accuracy on non-isomorphic pairs, but this result alone can be misleading. When identical graphs are presented with permuted node labels, LLMs fail to identify their isomorphism. Since permutation invariance is a fundamental requirement for valid structural reasoning, these results indicate that the success of LLMs on graph reasoning should not be strongly interpreted as evidence of genuine topological understanding.
\end{abstract}

\ccsdesc[500]{Computing methodologies~Natural language processing}
\ccsdesc[300]{Mathematics of computing~Graph theory}

\keywords{Large language models, graph isomorphism,
permutation invariance, graph reasoning}

\maketitle

\section{Introduction}

Large language models (LLMs) have demonstrated strong performance across a wide range of reasoning tasks, including mathematical problem-solving, proof-generation, and symbolic reasoning  \cite{wei2022emergent,li2026survey,kishanthan2026large}. This progress has motivated growing interest
in \textit{graph reasoning}, where a model must infer properties of graph entities
from their relations and the topology connecting them
\cite{amayuelas2025grounding,ke2025survey,ren2024survey}. Graph reasoning is
particularly challenging because structural information is distributed across
multiple nodes and edges and may require multi-step relational reasoning. Moreover, graphs are inherently non-sequential, whereas LLMs process
information as sequences of tokens. Consequently, strong accuracy on graph
tasks does not necessarily establish that an LLM understands the underlying
structure. It may instead exploit simple graph statistics, serialization cues,
or response biases.

Graph isomorphism problem   \cite{zemlyachenko1985graph} asks whether two graphs are structurally identical under a bijection between their node sets that preserves adjacency. Node identifiers are arbitrary and carry no structural meaning; therefore relabeling them does not change the graph’s underlying structure. Therefore, a strong graph reasoner should recognize the original graph and its relabeled copy as
isomorphic. This makes graph isomorphism a natural diagnostic for separating
surface-level difference detection from genuine structural comparison for graphs.

The Weisfeiler--Lehman (WL) hierarchy
\cite{weisfeiler1968reduction} is widely used to distinguish many
non-isomorphic graphs and characterize the expressive power of graph learning
methods \cite{huang2021short,xu2019powerful}. Graph neural networks (GNNs) \cite{gilmer2017neural,wu2020comprehensive} have become the predominant approach to representation learning on graph-structured data. They are generally designed in a way such that node representations are permutation equivariant and graph-level predictions are invariant to node order \cite{morris2019weisfeiler,hevapathige2026permutation}.
This property follows from permutation-invariant neighborhood aggregation and
graph-level pooling mechanisms.

LLMs, in contrast, receive graphs as token sequences, such as edge lists, edge
indices, or adjacency matrices. Their architectures do not explicitly enforce
invariance to node identifiers or serialization order. A relabeling can alter
the token sequence and attention pattern even though it leaves the graph
unchanged. Therefore, an LLM may correctly reject two visibly different graphs
yet fail to recognize two presentations of the same structure.

In this paper, we use a two-experiment protocol to report this gap through class-conditional performance. Experiment~1 evaluates non-isomorphic
graph pairs. The evaluated LLMs achieve near-perfect accuracy, which, when viewed
alone, appears to show strong graph-isomorphism detection ability. Experiment~2 then
presents graphs together with randomly permuted copies of themselves. The
same models usually declare these pairs non-isomorphic, despite their
isomorphism being guaranteed by construction.  Our contributions are as follows:

\begin{itemize}
    \item We report a class-conditional diagnostic that contrasts performance on
    non-isomorphic graph pairs with recognition of the same graph after random
    node permutation.

    \item We show that three widely used LLMs achieve near-perfect accuracy when rejecting non-isomorphic pairs, yet exhibit sharply lower sensitivity when asked to recognize permuted copies of the same graph.

    \item We identify several consistent behavioral outcomes: a strong
    preference for the \emph{non-isomorphic} label, substantial dependence on
    graph serialization, and only limited improvement from explicit
    instructions about permutation invariance.

    \item We argue that graph-reasoning evaluations should include permutation-based controls, because success at detecting
    differences does not establish understanding of structural equivalence.
\end{itemize}

\section{Related Work}

LLMs have demonstrated broad reasoning abilities
\cite{ning2024can,wu2025llms,xie2026core}, but studies also report their brittleness
on symbolic and systematic reasoning tasks
\cite{valmeekam2022large,li2024llms,agrawal2025can,zhang2025comprehension}.
Related robustness work also shows that semantically irrelevant perturbations can
change LLM model outputs in a considerable manner \cite{alahmari2025large,chen2026nlperturbator}. Recent studies have further shown that LLM predictions can depend on the order
of semantically equivalent inputs
\cite{tan2024order,guan2025order,egressy2025set,herbst2025lost}. For example,
permuting answer choices in a multiple-choice question can change predictions because of positional bias
under uncertainty \cite{pezeshkpour2024large}. This concern is particularly
important for graphs, where node identifiers, edge order, and serialization
syntax are inherently arbitrary.

IsoBench~\cite{fu2024isobench} evaluates graph isomorphism by using randomly permuted positive pairs and count-matched negative pairs. The findings indicate near-chance textual accuracy and default-answer behavior. Our contribution is the separate reporting of sensitivity and specificity, along with the use of degree-matched negative pairs and the comparison of three different textual serializations. This analysis explicitly shows how aggregate accuracy can mask a strong preference in responses. Additionally, Talk like a Graph~\cite{fatemi2024talk} demonstrates that the encoding of graphs impacts the performance of LLMs. Our serialization results specifically examine this relationship in the context of isomorphism judgments.

Herbst et al. ~\cite{herbst2025lost} investigate node labeling, edge encoding, and syntax in graph-reasoning tasks, reporting that model scale and fine-tuning enhance robustness. Our evaluation adds to their work by looking specifically at class-conditional isomorphism errors. 

\section{Experimental Setup}

\subsection{Preliminaries}

Let $G=(V,E)$ denote a finite, undirected, unweighted graph. Two graphs
$G_1=(V_1,E_1)$ and $G_2=(V_2,E_2)$ are \emph{isomorphic}, denoted
$G_1\cong G_2$, if there exists a bijection
$\phi:V_1\rightarrow V_2$ such that
\begin{equation}
(u,v)\in E_1 \iff (\phi(u),\phi(v))\in E_2.
\end{equation}

For any permutation $\pi$ of $V$, let $\pi(G)$ be the graph obtained by
replacing each node $v$ by $\pi(v)$ while preserving all adjacencies. By
construction,
\begin{equation}
G\cong \pi(G).
\end{equation}

\subsection{Large Language Models}

We evaluate three instruction-tuned LLMs from different providers:
\textbf{GPT-4o-mini} (OpenAI) \cite{achiam2023gpt},
\textbf{Gemini 2.5 Flash} (Google) \cite{team2023gemini}, and
\textbf{Llama 3.3 70B Instruct}, accessed through Together AI
\cite{touvron2023llama}. These models were selected because they are widely
used, accessible through hosted APIs, and representative of different model
families. Note that our results apply to the these evaluated models and settings.

\subsection{Joint Evaluation Protocol}

The non-isomorphic pairs from Experiment~1 and the permuted-isomorphic pairs
from Experiment~2 were evaluated as part of the same experimental run. For
each model, prompt, and serialization condition, both classes used the same
API endpoint, temperature, output labels, and parsing procedure. The task
instruction did not reveal the experiment, data-generation procedure, or true
class of a query. This joint protocol removes a run-level label cue: the model could not succeed
merely because all queries in one evaluation condition shared the same known
label. It also makes the comparison less vulnerable to changes in endpoint or
decoding configuration between experiments. All experiments use temperature zero to reduce decoding variation.

\subsection{Prompting Strategies}

We examine two prompting strategies. Neither condition explicitly asks for intermediate reasoning, and the instructed prompt requires only a label response.
\begin{itemize}
    \item \textbf{Zero-shot prompting:} The model receives the graph pair with only a minimal instruction (e.g., ``Are these two graphs isomorphic?'').

    \item \textbf{Instructed prompting:} The model receives a detailed prompt that explicitly defines graph isomorphism, clarifies that node identifiers are arbitrary and should be ignored, and highlights useful structural properties to guide reasoning. The complete prompt is provided below.
\end{itemize}

\begin{promptbox}{Instructed Prompt}
\small
You are given two finite, undirected, unweighted graphs.

\medskip
Two graphs are \textbf{isomorphic} if there exists a one-to-one mapping between
their nodes that preserves every edge and non-edge. Node identifiers are
arbitrary: different node names or presentation orders do not imply that the
graphs are different.

\medskip
Determine whether Graph~1 and Graph~2 are isomorphic. You may compare properties
such as the number of nodes, number of edges, degree sequence, neighbourhood
structure, connected components, cycles, and possible node correspondences.
Remember that matching simple statistics alone is not sufficient to prove
isomorphism.

\medskip
Return exactly one of the following labels and no additional text:

\begin{center}
\texttt{ISOMORPHIC}\\
\texttt{NON-ISOMORPHIC}
\end{center}

\textbf{Graph 1:}\\
\texttt{[GRAPH\_1]}

\medskip
\textbf{Graph 2:}\\
\texttt{[GRAPH\_2]}
\end{promptbox}

\subsection{Graph Serialization Methods}

We use three common graph serialization formats. An \textbf{edge list}
enumerates each connection as a node pair. An \textbf{edge index} stores
source and target node sequences, following the representation used in graph
learning frameworks \cite{fey2019fast}. An \textbf{adjacency matrix} encodes
the full binary connectivity matrix. A four-cycle can be represented as
follows:

\begin{table}[h]
\centering
\renewcommand{\arraystretch}{1.5}
\resizebox{0.625\columnwidth}{!}{%
\begin{tabular}{@{}ccc@{}}
\toprule
\textbf{Edge list} & \textbf{Edge index} & \textbf{Adjacency matrix} \\
\midrule
$0$--$1$ & $[0,1]$ & $0\;1\;0\;1$ \\
$0$--$3$ & $[0,3]$ & $1\;0\;1\;0$ \\
$1$--$2$ & $[1,2]$ & $0\;1\;0\;1$ \\
$2$--$3$ & $[2,3]$ & $1\;0\;1\;0$ \\
\bottomrule
\end{tabular}%
}
\end{table}

\vspace{-5mm}

These formats contain the same structural information but produce different
token sequences. Comparing them therefore tests whether the observed behavior
is stable across common textual representations of a graph.

\subsection{Evaluation metrics}

We take isomorphic as the positive class. Sensitivity (TPR) and specificity (TNR) are defined as
\begin{equation}
\mathrm{TPR}=\frac{\mathrm{TP}}{\mathrm{TP}+\mathrm{FN}}, \quad 
\mathrm{TNR}=\frac{\mathrm{TN}}{\mathrm{TN}+\mathrm{FP}}.
\end{equation}
where TP, FN, TN, and FP denote true positives, false negatives, true negatives, and false positives, respectively. We report specificity for the non-isomorphic cases and sensitivity for the permuted-isomorphic cases. Their equal-weight mean provides a balanced summary of the two class-conditional results.

\section{Two-Experimental Protocol}

\subsection{Experiment 1: Success on Non-Isomorphic Pairs}

\begin{figure}[htbp]
    \centering
    \includegraphics[width=0.6\linewidth]{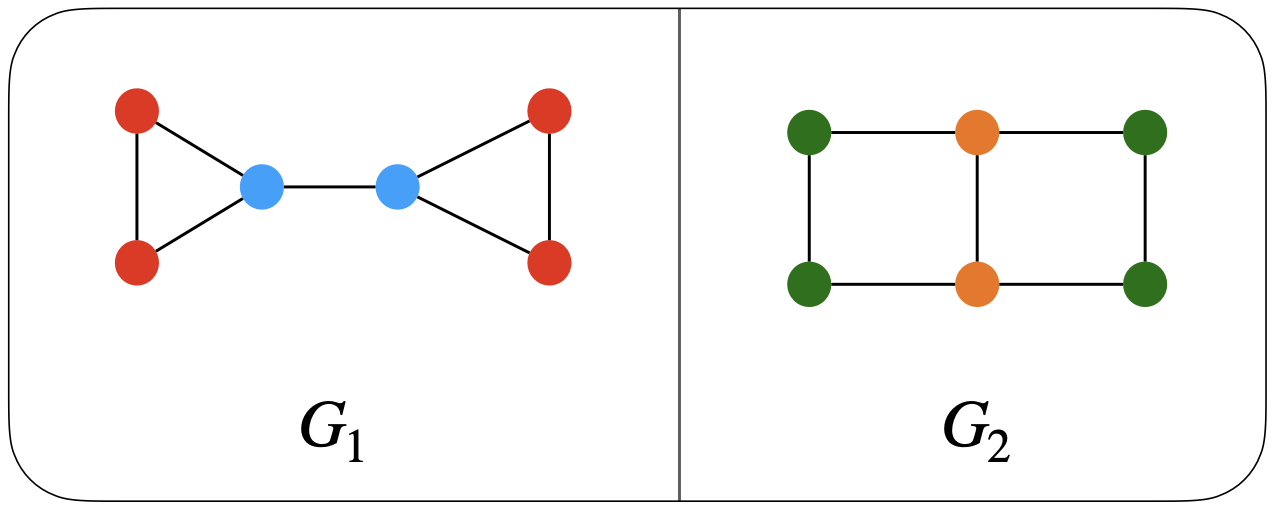}
    \caption{A non-isomorphic graph pair with the same degree sequence. The
    graphs differ structurally: $G_1$ contains triangles, whereas $G_2$
    contains 4-cycles.}
    \label{fig:graph-pair}
\end{figure}

The first experiment contains 400 non-isomorphic graph pairs with 8--15 nodes.
We construct four categories with 100 pairs each:

\begin{itemize}
    \item \textbf{Different nodes:} the graphs have equal edge counts but
    different node counts.

    \item \textbf{Different edges:} the graphs have equal node counts but
    different edge counts.

    \item \textbf{Different nodes and edges:} both graph size and edge count
    differ.

    \item \textbf{Degree matched:} the graphs have equal node counts, edge
    counts, and degree sequences, but are non-isomorphic. This removes the
    simple counting cues available in the first three categories (see Figure \ref{fig:graph-pair}).
\end{itemize}

\begin{table*}[t]
\centering
\small
\caption{Accuracy (specificity, \%) on the 400 non-isomorphic pairs.}
\label{tab:noniso-format}
\begin{tabular}{l|cc|cc|cc}
\hline
& \multicolumn{2}{c|}{OpenAI} & \multicolumn{2}{c|}{Gemini} & \multicolumn{2}{c}{Llama} \\
Format & Zero-shot & Instructed & Zero-shot & Instructed & Zero-shot & Instructed \\
\hline
Edge list        & 100.0 & 99.8 & 100.0 & 100.0 & 98.8 & 99.0 \\
Edge index       & 100.0 & 98.8 & 100.0 & 100.0 & 99.5 & 98.2 \\
Adjacency matrix & 100.0 & 91.5 & 100.0 & 100.0 & 97.5 & 100.0 \\
\hline
\end{tabular}
\end{table*}

\begin{table*}[t]
\centering
\small
\caption{Accuracy (specificity, \%) by non-isomorphic category, aggregated
over the three serialization formats.}
\label{tab:noniso-category}
\begin{tabular}{l|cc|cc|cc}
\hline
& \multicolumn{2}{c|}{OpenAI} & \multicolumn{2}{c|}{Gemini} & \multicolumn{2}{c}{Llama} \\
Category & Zero-shot & Instructed & Zero-shot & Instructed & Zero-shot & Instructed \\
\hline
Different nodes           & 100.0 & 100.0 & 100.0 & 100.0 & 99.7 & 99.3 \\
Different edges           & 100.0 & 98.7  & 100.0 & 100.0 & 96.0 & 98.3 \\
Different nodes and edges & 100.0 & 100.0 & 100.0 & 100.0 & 100.0 & 100.0 \\
Degree matched            & 100.0 & 88.0  & 100.0 & 100.0 & 98.7 & 98.7 \\
\hline
\end{tabular}
\end{table*}

\noindent\textbf{Results.} Table~\ref{tab:noniso-format} shows 91.5--100\% accuracy across every
model--prompt--format combination. Gemini correctly labels all evaluated
non-isomorphic pairs. OpenAI and Llama are also near perfect. The same pattern
holds across graph categories in Table~\ref{tab:noniso-category}. Notably, the
models obtain 88.0--100\% accuracy on degree-matched pairs, where node count,
edge count, and degree sequence cannot distinguish the graphs.

High specificity alone does not establish graph isomorphism competence. For example, a predictor that always labels graphs as non-isomorphic could still achieve a 100\% accuracy in this experiment, including within the degree-matched category. However, true graph isomorphism requires both the ability to differentiate between different structures and the capability to recognize identical structures. Experiment 2 directly tests this second requirement.

\subsection{Experiment 2: Failure on Permuted Copies of the Same Graph}

\begin{figure}[htbp]
    \centering
    \includegraphics[width=0.525\columnwidth]{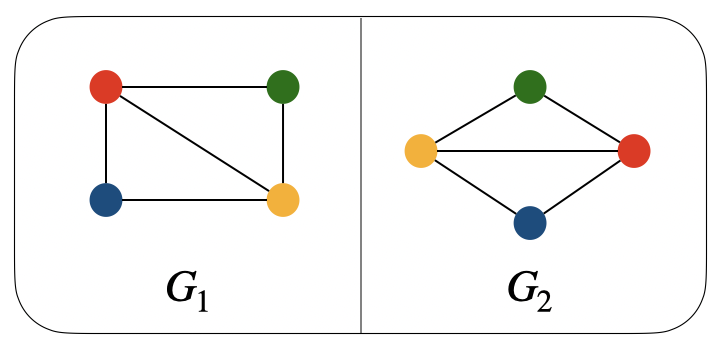}
    \caption{Two isomorphic graphs with different node labelings and visual layouts. Despite these representational differences, \(G_1\) and \(G_2\) have the same underlying graph structure.}
    \label{fig:isomorphic_graph_pair}
\end{figure}

We generate 400 random connected graphs, with 100 graphs for each size
$n\in\{5,10,15,20\}$. For every graph $G$, we sample a random permutation
$\pi$ of its node set and construct the pair $(G,\pi(G))$ (see Figure \ref{fig:isomorphic_graph_pair}). The second graph is
therefore not merely similar to the first: it is the same graph with different
node identifiers. Every pair is isomorphic by construction.

\begin{table}[t]
\centering
\footnotesize
\setlength{\tabcolsep}{3pt}
\caption{Sensitivity on permuted copies of the same graph, with 95\% Wilson
score confidence intervals.}
\label{tab:permutation}
\begin{tabular}{llccc}
\hline
\textbf{LLM} & \textbf{Prompt} & \textbf{Serialization} & \textbf{Sensitivity} & \textbf{95\% CI} \\
\hline
\multirow{6}{*}{OpenAI}
& \multirow{3}{*}{Zero-shot}
    & Edge list       & 0.0\%  & [0.000, 0.010] \\
&   & Edge index      & 0.0\%  & [0.000, 0.010] \\
&   & Adjacency matrix& 0.5\%  & [0.001, 0.018] \\
\cline{2-5}
& \multirow{3}{*}{Instructed}
    & Edge list       & 11.2\% & [0.085, 0.147] \\
&   & Edge index      & 16.0\% & [0.127, 0.199] \\
&   & Adjacency matrix& 18.0\% & [0.145, 0.221] \\
\hline
\multirow{6}{*}{Gemini}
& \multirow{3}{*}{Zero-shot}
    & Edge list       & 0.2\%  & [0.000, 0.014] \\
&   & Edge index      & 0.2\%  & [0.000, 0.014] \\
&   & Adjacency matrix& 0.0\%  & [0.000, 0.010] \\
\cline{2-5}
& \multirow{3}{*}{Instructed}
    & Edge list       & 0.0\%  & [0.000, 0.010] \\
&   & Edge index      & 0.0\%  & [0.000, 0.010] \\
&   & Adjacency matrix& 0.0\%  & [0.000, 0.010] \\
\hline
\multirow{6}{*}{Llama}
& \multirow{3}{*}{Zero-shot}
    & Edge list       & 11.2\% & [0.085, 0.147] \\
&   & Edge index      & 6.0\%  & [0.041, 0.088] \\
&   & Adjacency matrix& 2.0\%  & [0.010, 0.039] \\
\cline{2-5}
& \multirow{3}{*}{Instructed}
    & Edge list       & 31.5\% & [0.271, 0.362] \\
&   & Edge index      & 39.2\% & [0.346, 0.441] \\
&   & Adjacency matrix& 4.5\%  & [0.029, 0.070] \\
\hline
\end{tabular}
\end{table}
\vspace{\baselineskip}

\noindent\textbf{Results.} Table~\ref{tab:permutation} reverses the impression created by Experiment~1.
OpenAI achieves only 0--18\% sensitivity, Gemini 0--0.2\%, and Llama
2--39.2\%. Thus, in every evaluated condition, most graphs are judged
non-isomorphic to permuted copies of themselves. The failure is especially striking for Gemini. The model is perfect on all
non-isomorphic conditions in Experiment~1 but almost never recognizes an
isomorphic pair in Experiment~2. OpenAI exhibits the same pattern in the
zero-shot setting: near-perfect rejection of non-isomorphic pairs but almost
zero recognition of permuted copies. Llama is less extreme, particularly with
the instructed edge-index prompt, but even its best sensitivity is only
39.2\%.

Explicit instructions help OpenAI and Llama in several conditions. For
example, Llama improves from 6.0\% to 39.2\% on edge indices. Nevertheless,
the instructed prompt explicitly states that node identifiers are arbitrary,
yet most predictions remain incorrect. Gemini receives no measurable benefit
from the additional instruction. Serialization also matters. Under instructed prompting, Llama ranges from
39.2\% with edge indices to 4.5\% with adjacency matrices. Since these formats
encode the same graph, this variation indicates that the models' judgments
depend strongly on how structure is converted into tokens.

\section{Combined Analysis}

\subsection{Balanced Performance}

To summarize performance across the two true classes, we average specificity
from Experiment~1 and sensitivity from Experiment~2 with equal weight:
\begin{equation}
\mathrm{Balanced\ Accuracy}
=\frac{\mathrm{TNR}+\mathrm{TPR}}{2}.
\end{equation}

\begin{table}[t]
\centering
\footnotesize
\setlength{\tabcolsep}{3.5pt}
\caption{Equal-class balanced accuracy (\%) obtained by combining the
specificity and sensitivity reported in the two experiments.}
\label{tab:balanced}
\begin{tabular}{llccc}
\toprule
\textbf{LLM} & \textbf{Prompt} & \textbf{Edge list} & \textbf{Edge index} & \textbf{Adjacency matrix} \\
\midrule
\multirow{2}{*}{OpenAI}
& Zero-shot  & 50.0 & 50.0 & 50.3 \\
& Instructed & 55.5 & 57.4 & 54.8 \\
\midrule
\multirow{2}{*}{Gemini}
& Zero-shot  & 50.1 & 50.1 & 50.0 \\
& Instructed & 50.0 & 50.0 & 50.0 \\
\midrule
\multirow{2}{*}{Llama}
& Zero-shot  & 55.0 & 52.8 & 49.8 \\
& Instructed & 65.3 & 68.7 & 52.3 \\
\bottomrule
\end{tabular}
\end{table}

As Table~\ref{tab:balanced} shows, the resulting values range from 49.8\% to
68.7\%, far below the near-perfect results suggested by Experiment~1 alone.
Gemini's balanced accuracy is approximately 50\% because it nearly always
selects the non-isomorphic label. Llama with instructed edge indices achieves
the strongest balance, but still fails on most isomorphic cases.

\vspace{-4mm}

Because the two classes use different graph-generation procedures and partly
different size ranges, these values should be interpreted as an equal-weight
summary of the reported class-conditional rates rather than as accuracy on a
perfectly distribution-matched benchmark. This limitation does not affect the
central observation that every pair in Experiment~2 is provably isomorphic and
is nevertheless rejected in most cases.

\subsection{What the Two Experiments Show}

The combined evidence supports four main conclusions.

\noindent\textbf{Detecting difference $\neq$ understanding isomorphism.} 
Models perform well on rejecting different graphs (Experiment~1) but poorly 
on recognizing structural identity despite arbitrary names (Experiment~2). 
Success on the former does not imply general isomorphism competence.

\noindent\textbf{Strong response asymmetry.} High specificity and low sensitivity indicate a bias toward predicting non-isomorphic relationships. Testing both classes under identical conditions removes a run-level label cue, but it does not eliminate a learned response preference.

\noindent\textbf{Low recognition of relabeled copies.} Permutations 
preserve all structural properties (adjacency, paths, cycles, etc.). 
Most $(G,\pi(G))$ pairs are rejected, but the missing $(G,G)$ control prevents attributing this specifically to relabeling.

\noindent\textbf{Prompting helps but does not guarantee invariance.} Explicit 
isomorphism definitions and warnings about arbitrary names improve some 
OpenAI/Llama conditions, but gains are inconsistent and insufficient. 
Natural language lacks the permutation-invariance guarantee of conventional 
graph models.

\vspace{-4mm}

\section{Conclusion}

The evaluated models demonstrate high specificity but low sensitivity when it comes to graph isomorphism. This gap shows why evaluations should report sensitivity and specificity separately. While explicit instructions improve several conditions, they do not completely eliminate errors. These experiments do not indicate an architectural limitation or dismiss the potential benefits of reasoning prompts or additional training. For graph-augmented LLM pipelines the practical implication is to test predictions across equivalent graph serializations and node relabelings. Implementing canonical labeling before prompting is one potential safeguard.

\noindent\textbf{Limitations.} We consider three representative LLMs and three serialization methods. Although these settings reveal consistent permutation sensitivity, they do not cover the wide range of available models and input representations. The graph instances were generated randomly without explicitly controlling for structural properties such as girth, diameter, symmetry, density, or automorphism structure. Consequently, some graph classes may be easier or more difficult for LLMs than others. We also did not systematically isolate the effects of tokenization, token embeddings, and context length. We did not evaluate identical-input pairs \((G,G)\) or report sensitivity separately by graph size. Therefore, we are unable to distinguish between response preferences and the effects of relabeling, nor can we determine how sensitivity varies with graph size. Confidence intervals are provided only for the primary sensitivity results. Additionally, no chain-of-thought condition or specific reasoning configuration was tested. Finally, our experiments cannot determine whether the observed failures arise from limitations in symbolic reasoning, sensitivity to sequential input representations, or reliance on statistical patterns learned during pretraining.

\noindent\textbf{Future Work.} We plan to evaluate a broader range of models, prompting strategies, graph sizes, and serialization methods. Controlled graph generators could be used to examine how structural properties affect accuracy and permutation invariance. It would be useful to evaluate approaches such as graph-aware tokenization and canonical labeling, alongside specialized prompting and fine-tuning. Mechanistic analyses of internal representations may help explain why equivalent graph inputs produce different predictions. Finally, similar invariance tests could be extended to other domains in which input order should not affect the answer, including set reasoning, database querying, and knowledge graph reasoning.

\begin{acks}
The authors used AI-assisted tools for language polishing and improvements to
readability and clarity of the manuscript. All AI-assisted edits were reviewed and approved by
the authors.
\end{acks}

\bibliographystyle{ACM-Reference-Format}
\bibliography{custom}

\end{document}